\documentclass[runningheads]{llncs}

\usepackage{eccv}

\usepackage{eccvabbrv}

\usepackage{graphicx}
\usepackage{booktabs}

\usepackage[accsupp]{axessibility}  

\usepackage{wrapfig}
\usepackage{multirow}
\usepackage{marvosym}

\usepackage[pagebackref,breaklinks,colorlinks,citecolor=eccvblue]{hyperref}

\usepackage{orcidlink}

\begin{document}

\title{Sparse-View Surface Reconstruction using Gaussian Splatting through High-Confidence Depth Propagation with Normal Priors} 

\titlerunning{Sparse-View Surface Reconstruction}

\author{Liang Han\inst{1} \and
Bangcai Wei\inst{2} \and
Junsheng Zhou\inst{1}\textsuperscript{\Letter} \and
Yu-Shen Liu\inst{1}\textsuperscript{\Letter} \and
Zhizhong Han\inst{3}
}

\authorrunning{L.~Han et al.}

\institute{School of Software, Tsinghua University, Beijing, China \\
\email{\{hanl23, zhou-js24\}@mails.tsinghua.edu.cn, \\ liuyushen@tsinghua.edu.cn}
\and
China Telecom
\email{weibc.gs@chinatelecom.cn}
\and
Department of Computer Science, Wayne State University, Detroit, USA
\email{h312h@wayne.edu}
}

\maketitle

\begin{abstract}
3D reconstruction from sparse views is a challenging task in 3D computer vision. Recent studies on 3D Gaussian Splatting (3DGS) have achieved remarkable results with sparse views in novel view synthesis, yet reconstructing high-quality geometric surfaces from sparse views remains a challenge, due to the limited geometry clues and the discreteness of Gaussians.
In this paper, we propose a novel 3DGS-based method for high-fidelity surface reconstruction from sparse views. Our key insight is to introduce a normal-guided depth propagation approach, which can extend depth information from high-confidence regions to constrain the depth in low-confidence areas. Additionally, we propose an abnormal depth edge-aware regularization to address depth discontinuities caused by the discreteness of Gaussians.
Extensive experiments on DTU and Tanks-and-Temples datasets demonstrate that our method outperforms the state-of-the-art methods in sparse view surface reconstruction. Project page: \url{https://hanl2010.github.io/DP-GS}.
\end{abstract}

\section{Introduction}
\label{sec:intro}

3D reconstruction technology has advanced rapidly with the development of deep learning. The emergence of neural radiance field \cite{2021nerf} has significantly improved both of the reconstruction quality and efficiency in surface reconstruction tasks  \cite{wang2021neus, yariv2021volsdf, wang2022hfneus, fu2022geoneus, darmon2022neuralwarp, yu2022monosdf, wang2023neus2, wu2022voxurf, li2023neuralangelo}. More recently, the emergence of 3D Gaussian Splatting (3DGS)~\cite{3dgs} has further improved rendering efficiency, accelerating rendering-based reconstruction methods \cite{guedon2023sugar, dai2024gs_surfel, huang20242dgs, lyu20243dgsr, yu2024gof, pgsr, zhang2025gspull, wu2024adv_3dgs}. However, these methods typically require a dense set of input views. With limited number of input views, their performance degenerates significantly, severely degrading the reconstruction quality.

\begin{figure}
    \centering
    \includegraphics[width=1.0\linewidth]{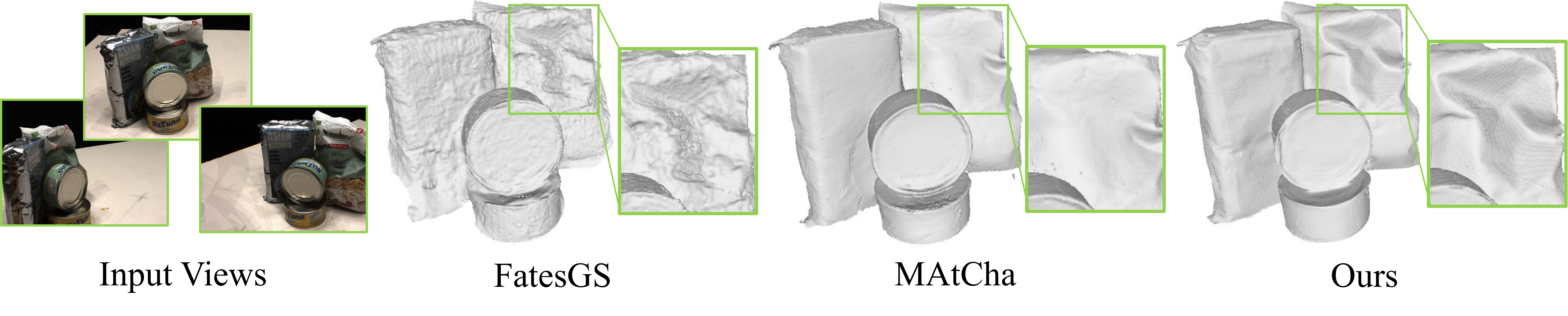}
    \caption{Reconstructed surfaces from three small-overlap views of the DTU dataset. Compared with state-of-the-art 3DGS-based methods FatesGS \cite{fatesgs} and MAtCha \cite{guedon2025matcha}, Our method produces more faithful surface reconstruction.}
    \label{fig:cover_fig}
\end{figure}

Two types of solutions have been proposed for addressing this issue: generalizable methods \cite{long2022sparseneus, ren2023volrecon, peng2023gens, xu2023c2f2neus, liang2024retr} and scene-specific optimization methods \cite{yu2022monosdf, wu2023svolsdf, huang2023neusurf, sparsecraft, fatesgs, han2025sparserecon}. Generalizable methods require data-driven training on large-scale datasets to generalize learned priors to new scenes.
However, the pre-training process is time-consuming, and the reconstruction degenerates when the input views do not match the sparse view configuration used during training.
Thus, they usually struggle to reconstruct surfaces when testing scenes differ from training scenes a lot.

In contrast, scene-specific optimization methods do not require training on large-scale datasets but instead fitting 3D geometry directly from sparse views. However, some current scene-specific sparse view reconstruction methods \cite{wu2023svolsdf, huang2023neusurf, han2025sparserecon} adopt NeRF \cite{2021nerf} as the representation, which is slow to optimize. Compared to NeRF \cite{2021nerf}, 3DGS \cite{3dgs} significantly improves the rendering efficiency and quality. Although numerous studies \cite{li2024dngaussian, zhu2025fsgs, han2024binocular, chen2024mvsplat, zhang2024corgs, fatesgs, wu2025sparse2dgs} have explored 3DGS-based approaches under sparse-view conditions, most of them focused on novel view synthesis rather than reconstructing the geometry. 
Due to the limited geometry clue and the discreteness of Gaussians, it is susceptible to ambiguous Gaussian primitives when processing sparse input views, leading to defective depth rendering and the failure of surface reconstruction.

In this paper, we propose a novel 3DGS-based scene-specific optimization method 
for high fidelity surface reconstruction from sparse views. To address the depth uncertainty, we introduce a depth propagation technique that effectively supervises ambiguous regions. Specifically, we leverage both multi-view geometric and photometric consistency to identify regions with high-confidence depth which is more reliable during the inference.
We then utilize the normal prior to propagate these high-confidence depths along the direction perpendicular to the normal. This approach enables robust supervisions on regions with low-confidence depth, ultimately leading to more accurate geometric surface reconstruction.

However, the accuracy of the propagated depths decreases as the propagation range increases, making it challenging to effectively supervise occluded or low-texture regions far from area with high-confidence depth. Moreover, in these regions, depth confidence cannot be determined through multi-view consistency. 
These facts make some areas may exhibit depth discontinuities during Gaussian splitting optimizing, especially when only using the normal prior to impose constraints. To maintain depth continuity in the under-constrained regions, we propose an edge-aware smoothing loss for abnormal depth, which mitigates excessive Gaussian discretization and promotes smoother local surfaces, ultimately improving surface reconstruction quality.

In summary, our main contributions are as follows:
\begin{itemize}
    \item We propose a novel 3DGS-based method for high-fidelity surface reconstruction from sparse views. The key idea lies in a normal-guided depth propagation constraint which effectively supervises the learning of 3D Gaussians with uncertain depths.
    \item We introduce an edge-aware smoothing loss for abnormal depth to mitigate depth discontinuities caused by excessive Gaussian dispersion in under-constrained regions.
    \item Extensive experiments on widely-used forward-facing and 360-degree scene datasets demonstrate that our method achieves state-of-the-art results compared to existing sparse view surface reconstruction methods.
\end{itemize}

\begin{figure*}[t]
  \centering
   \includegraphics[width=0.95\textwidth]{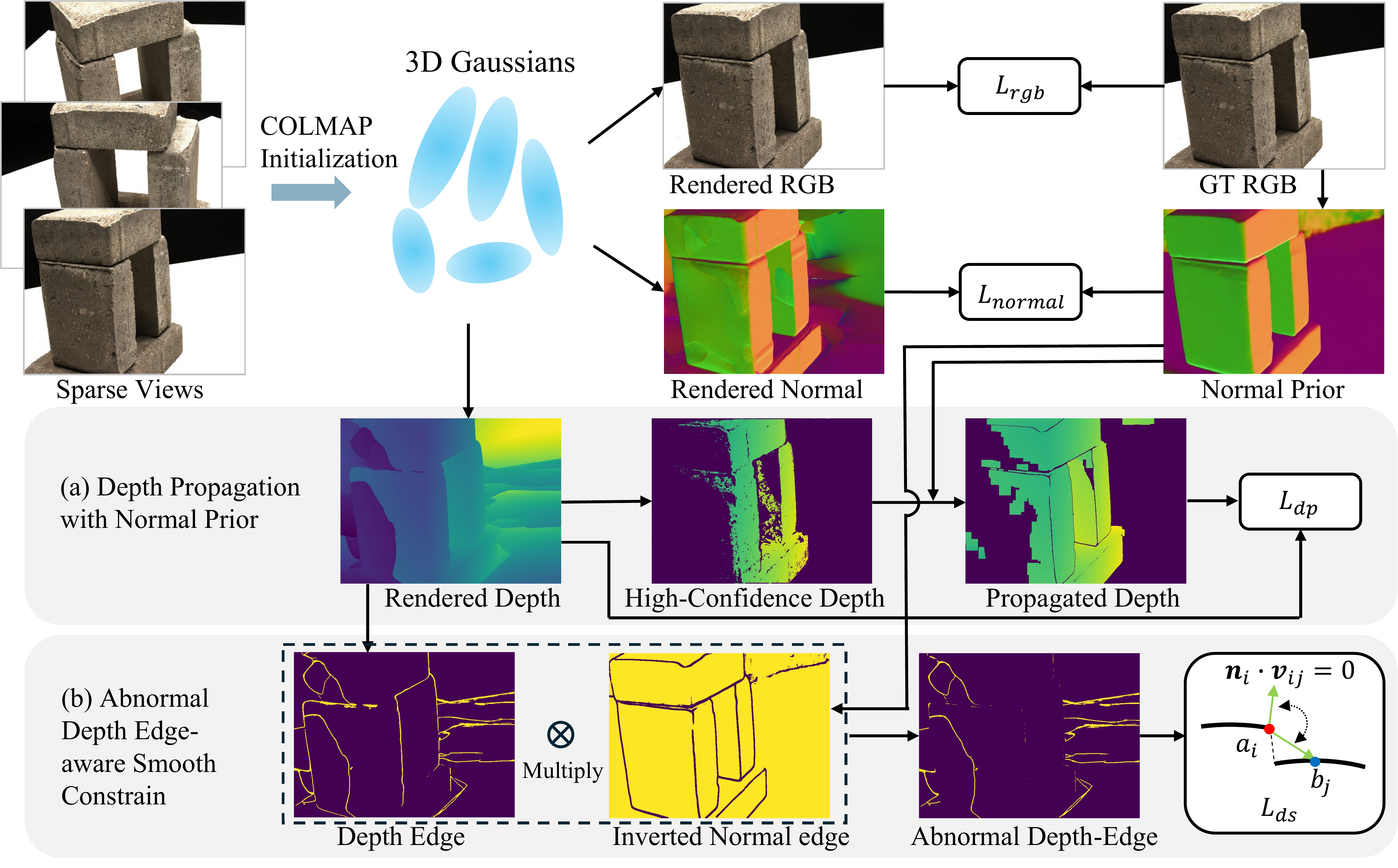}
   \caption{The overview of our method. (a) We leverage a normal prior to propagate depth information from high-confidence regions to low-confidence regions, effectively constraining the depth in regions with lower confidence. (b) The edge-aware regularization for abnormal depths is employed to eliminate depth discontinuities caused by the insufficient constraints of the unstructured Gaussian. In addition, we also use color loss and normal prior loss, as similar previous methods \cite{shen2024solidgs, wang2024gaussurf}.}
   \label{fig:pipeline}
\end{figure*}

\section{Related Work}
\label{sec:related}
\subsection{Neural Radiance Field}
Neural implicit representations have become a powerful paradigm for 3D scene understanding and reconstruction from both images \cite{2021nerf, barron2022mip360, instantngp, zhang2025nerfprior, zhang2025monoinstance, zhang2026vrp-udf} and point clouds \cite{yang2025swin3d, wang2024noise4denoise, li2025point_mask_trans, mao2025pcac, li2025learning-normals, li2025pff-net, wen2022pmp, xiang2022snowflake}.
The emergence of NeRF \cite{2021nerf} has led to the rapid development of neural volume rendering-based reconstruction methods. NeuS \cite{wang2021neus}, VolSDF \cite{yariv2021volsdf}, and other subsequent methods \cite{wang2022hfneus, fu2022geoneus, darmon2022neuralwarp, li2023neuralangelo, wu2022voxurf} introduced Signed Distance Functions (SDF) in NeRF as an implicit representation of 3D geometry to achieve multi-view 3D reconstruction. To improve the reconstruction efficiency, researchers have proposed a series of high-fidelity and computationally efficient methods, such as improvements in sampling strategies \cite{zhang2023shootingfewer, sun20223reconwild, dogaru2023sphere} or the use of hash grids \cite{wang2023neus2, cai2023neuda}.

Recently, 3DGS \cite{3dgs} has achieved unprecedented optimization speed and rendering quality in novel view synthesis tasks. However, it is challenging to extract smooth surfaces due to the discreteness and disorderliness of Gaussian distributions. Subsequent works have addressed this by flattening Gaussian ellipsoids \cite{guedon2023sugar, huang20242dgs, dai2024gs_surfel, li2026va-gs, li2025gaussianudf}, combining with neural implicit radiance fields \cite{lyu20243dgsr, yu2024gsdf, zhang2025gspull}, or improving the accuracy of rendered depth \cite{pgsr, zhang2024radegs} to obtain more stable and accurate geometric surfaces. 

Although these methods show significant quality and speed improvements in surface reconstruction, they heavily rely on dense views during the optimization process.

\subsection{Surface Reconstruction with Sparse Views}
Although there have been some methods based on NeRF or 3DGS with sparse view as input, most of them \cite{niemeyer2022regnerf, yang2023freenerf, jain2021dietnerf, deng2022dsnerf, wang2023sparsenerf, truong2023sparf, yuan2022rgbdnerf, li2024dngaussian, zhu2025fsgs, han2024binocular, zhang2024corgs, chen2024mvsplat} focus on novel view synthesis rather than geometric surface reconstruction. Recent studies on sparse view surface reconstruction can be divided into two categories: generalization based methods \cite{long2022sparseneus, ren2023volrecon, peng2023gens, xu2023c2f2neus, liang2024retr, na2024uforecon} or scene-specific optimization based methods \cite{yu2022monosdf, wu2023svolsdf, huang2023neusurf, fatesgs, sparsecraft, shen2024solidgs, wu2025sparse2dgs, han2025sparserecon}. Generalizable methods require pre-training on large-scale datasets, which is typically time-intensive, and the reconstruction results are often exhibit limited fidelity on unseen scenes. Additionally, reconstruction results can only be plausible when the sparse view configuration aligns with the pre-training setup. When there are significant differences in view angle setup, the reconstruction quality will degrade significantly. In particular, geometric reconstruction cannot be achieved effectively, when there are significant domain discrepancies between the training and testing data.

In contrast, scene-specific optimization methods do not require training on large-scale datasets but instead leverage monocular geometric cues \cite{ke2024marigold, hu2024metric3d, yang2024depthanything} to fit 3D geometry directly from sparse views. 
Some previous studies \cite{wu2023svolsdf, huang2023neusurf, han2025sparserecon} introduced prior constraints based on neural implicit reconstruction methods to achieve high-quality sparse view reconstruction, but they were limited by the slow convergence and heavy computational cost of implicit field optimization. SparseCraft \cite{sparsecraft} builds upon Instant-NGP \cite{instantngp} by incorporating multi-view stereo (MVS) priors to achieve fast sparse-view reconstruction. However, it is
only effective for sparse views with large overlap, and the reconstruction quality degrades significantly when the overlap of views decreases. 
Recently, FatesGS \cite{fatesgs}, SolidGS \cite{shen2024solidgs} and MAtCha \cite{guedon2025matcha} have achieved sparse view reconstruction based on 3DGS \cite{3dgs}, enabling fast optimization within minutes. Nevertheless, these methods still face challenges in producing smooth and complete surfaces, as illustrated in Fig.~\ref{fig:cover_fig}.

Although many multi-view reconstruction methods, such as SolidGS \cite{shen2024solidgs}, GaussianRoom \cite{xiang2024gaussianroom}, MonoSDF \cite{yu2022monosdf}, NeuRIS \cite{wang2022neuris} and GausSurf \cite{wang2024gaussurf}, leverage monocular normal priors to achieve higher-quality geometric reconstructions, they merely apply regularization to the rendered normal without fully exploiting the potential of normal priors. On the other hand, some monocular depth estimation methods, such as GeoNet+ \cite{qi2020geonet++} and IronDepth \cite{bae2022irondepth}, refine depth estimation accuracy by utilizing estimated normal. However, due to the distortions and scale ambiguity inherent in monocular depth estimation, these methods cannot effectively leverage normal priors and are not directly applicable for multi-view geometric reconstruction constraints.

\section{Methods}
\label{sec:method}
The pipeline of our method is illustrated in Fig.~\ref{fig:pipeline}. In this section, we first describe the process of obtaining the depth confidence map and propagating high-confidence depth using the normal prior. Next, we introduce the method of detecting abnormal depth edges and the smoothing loss. Finally, we present the overall loss function.

\subsection{Depth Propagation with Normal Prior}
The key to improving the quality of sparse view reconstruction is to ensure that 3DGS renders accurate and consistent depth across different views. However, due to the unstructured and discrete nature of Gaussians, relying solely on color reconstruction constraints on the input sparse views is insufficient to accurately distribute most Gaussians on the scene surface and obtain precise depth. A straightforward approach to incorporating additional supervision is to leverage monocular depth priors to constrain the depth rendered by 3DGS for sparse view reconstruction.  However, monocular depth priors are often distorted and fail to accurately align with the real scene surface even after calibration. 

To overcome the above difficulties, we propose a novel \textit{depth propagation constraint} that indirectly leverages the normal prior to constrain the depth rendered by 3DGS, where the normal prior is a image-based prior predicted by a pretrained normal estimation model. First, we jointly use multi-view geometric consistency and multi-view photometric consistency to predict a depth confidence map. Then, using the normal prior as guidance, we propagate high-confidence depths outward to constrain the regions with low depth confidence.

\begin{wrapfigure}{r}{0.5\linewidth}
    \centering
    \includegraphics[width=\linewidth]{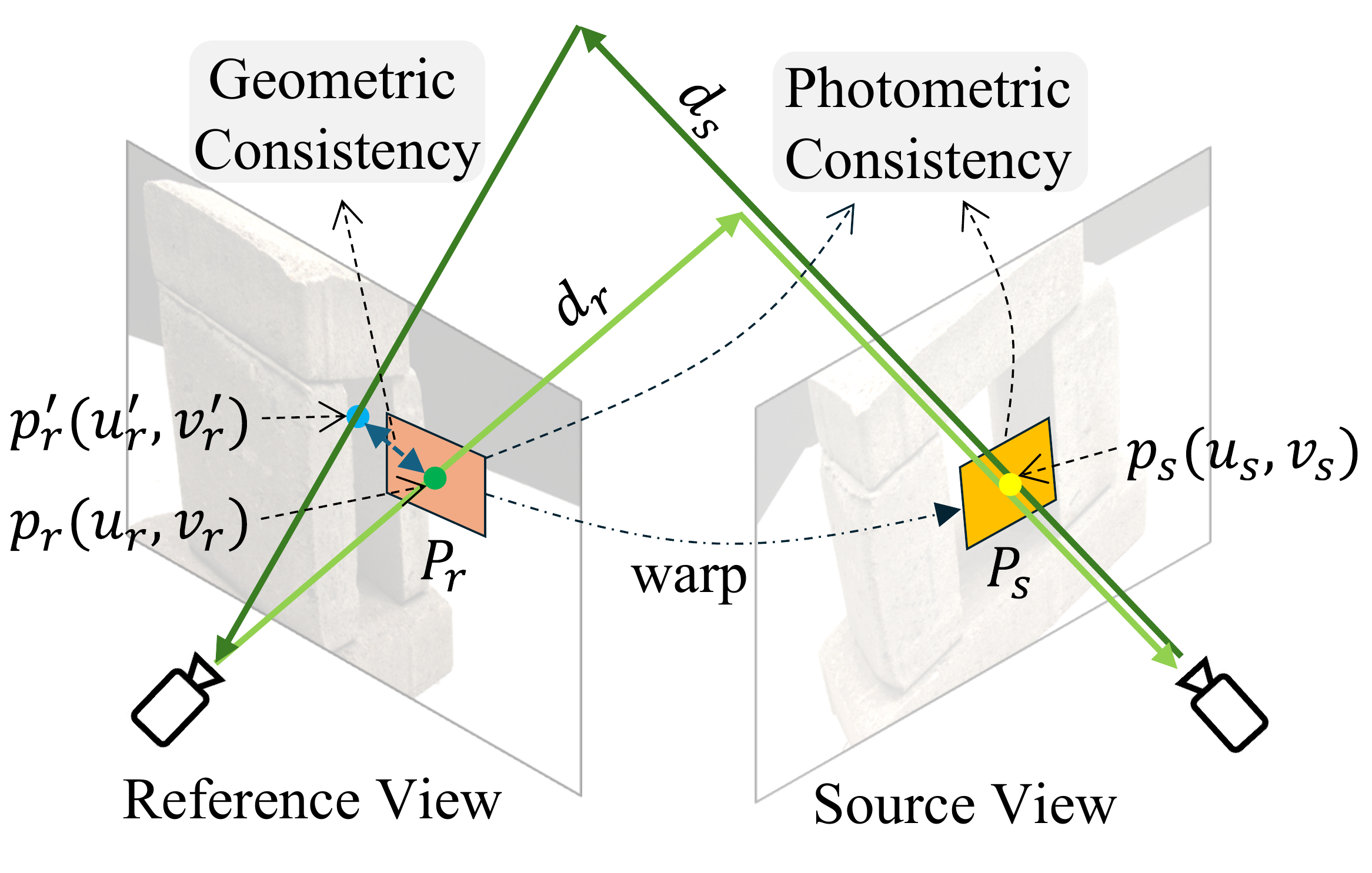}
    \caption{Illustration of depth confidence calculation.}
    \label{fig:depth-confidence}
\end{wrapfigure}

\subsubsection{Depth Confidence Map}
As shown in Fig.~\ref{fig:depth-confidence}, given two adjacent input views as the reference view $I_r$ and the source view $I_s$, their depth maps $D_r$ and $D_s$ can be obtained by rendering 3D Gaussians. For a particular pixel $p_r$ on the reference view with coordinates $(u_r, v_r)$ and its depth $d_r$ on $D_r$, it can be mapped to the source view through the depth-aware reprojection transformation $H_{rs}$, resulting in a projected pixel $p_s$ with coordinates $(u_s, v_s)$. Similarly, we can map the pixel $p_s(u_s, v_s)$ from the source view back to the reference view using the depth $d_s$ interpolated from $D_s$ and the depth-aware reprojection transformation $H_{sr}$, resulting in a pixel $p^{\prime}_r$ with coordinates $(u^{\prime}_r, v^{\prime}_r)$. The distance error between $p_r(u_r, v_r)$ and $p^{\prime}_r(u^{\prime}_r, v^{\prime}_r)$ reflects the consistency of the rendered depths $d_r$ and $d_s$, so we incorporate it as part of the depth confidence. The confidence map of geometric consistency is defined as
\begin{equation}
    C_{geo} = 
    \begin{cases}
        \frac{1}{e^{\left\| p_{r} - p^{\prime}_{r} \right\| }}, &\text{ if } \left\| p_{r} - p^{\prime}_{r} \right\| \leqslant 1 \\
        0, &\text{ if } \left\| p_{r} - p^{\prime}_{r} \right\| >1
    \end{cases}
    ,
\end{equation}

\begin{equation}
    p_s = H_{rs} p_r ,
\end{equation}

\begin{equation}
    p^{\prime}_r = H_{sr} p_s .
\end{equation}




Nevertheless, multi-view geometric consistency cannot fully reflect the accuracy of depth predictions, so we also leverage the patch-based multi-view photometric consistency to further determine the depth confidence. Specifically, as shown in Fig.~\ref{fig:depth-confidence}, for a patch $P_r$ centered at pixel $p_r$, we warp it to the source view using the homography matrix $H_{rs}$, resulting in a patch $\hat{P}_s$. We then employ Normalized Cross-Correlation (NCC) \cite{yoo2009ncc} to measure the photometric consistency between patch $P_r$ and $P_s$ as another part of the depth confidence as:
\begin{equation}
    C_{pho} = NCC(P_r, P_s) ,
\end{equation}
where $P_s = H_{rs}P_r $. Ultimately, we use the product of multi-view geometric consistency and multi-view photometric consistency to describe the depth confidence as:
\begin{equation}
    C = C_{geo} \cdot C_{pho} .
\end{equation}

It is noteworthy that, unlike PGSR \cite{pgsr}, we do not directly use multi-view geometric consistency and multi-view photometric consistency as loss terms to constrain the geometric surface. Instead, we primarily exploits them to estimate depth confidence, enabling the identification of regions with high-fidelity depth. By focusing on high-confidence depth estimation rather than direct multi-view optimization, our method mitigates the limitations of geometric consistency losses in sparse-view settings and leads to more robust reconstruction results.

\begin{wrapfigure}{r}{0.5\linewidth}
    \centering
    \includegraphics[width=\linewidth]{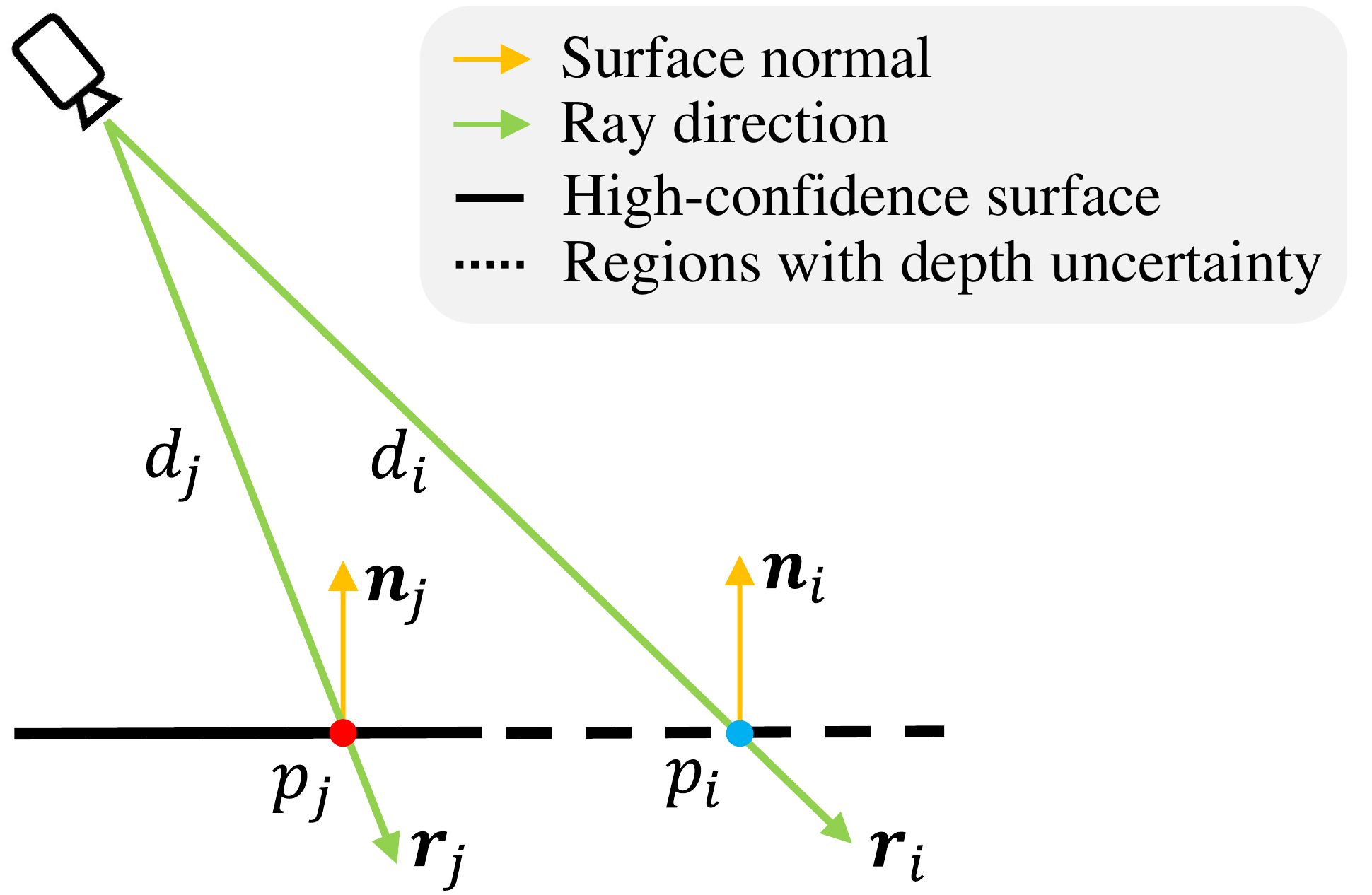}
    \caption{Illustration of the depth propagation.}
    \label{fig:depth-propagation}
\end{wrapfigure}

\subsubsection{Depth Propagation}
Our goal of depth propagation is to leverage normal priors to obtain a broader range of accurate depth predictions on planes, which contributes to eliminating the uncertainty of Gaussian depth at low-confidence regions. Fig.~\ref{fig:depth-propagation} illustrates the depth propagation algorithm in a local plane. Given a pixel $p_i$ with coordinates $(u_i,v_i)$ and a neighboring pixel $p_j$ with coordinates $(u_j,v_j)$ and depth $d_j$, we define the normalized ray directions from the camera center passing through these two pixels as $\boldsymbol{r}_i$ and $\boldsymbol{r}_j$, respectively. If pixel $p_i$ and its neighboring pixel $p_j$ belong to the same plane, meaning their normals are identical, then the depth of pixel $p_i$ is given by:
\begin{equation}
    d_i = \frac{\boldsymbol{n}_j \boldsymbol{r}_j}{\boldsymbol{n}_i \boldsymbol{r}_i} d_j.
\end{equation}

\begin{figure}[b]
    \centering
    \includegraphics[width=0.8\linewidth]{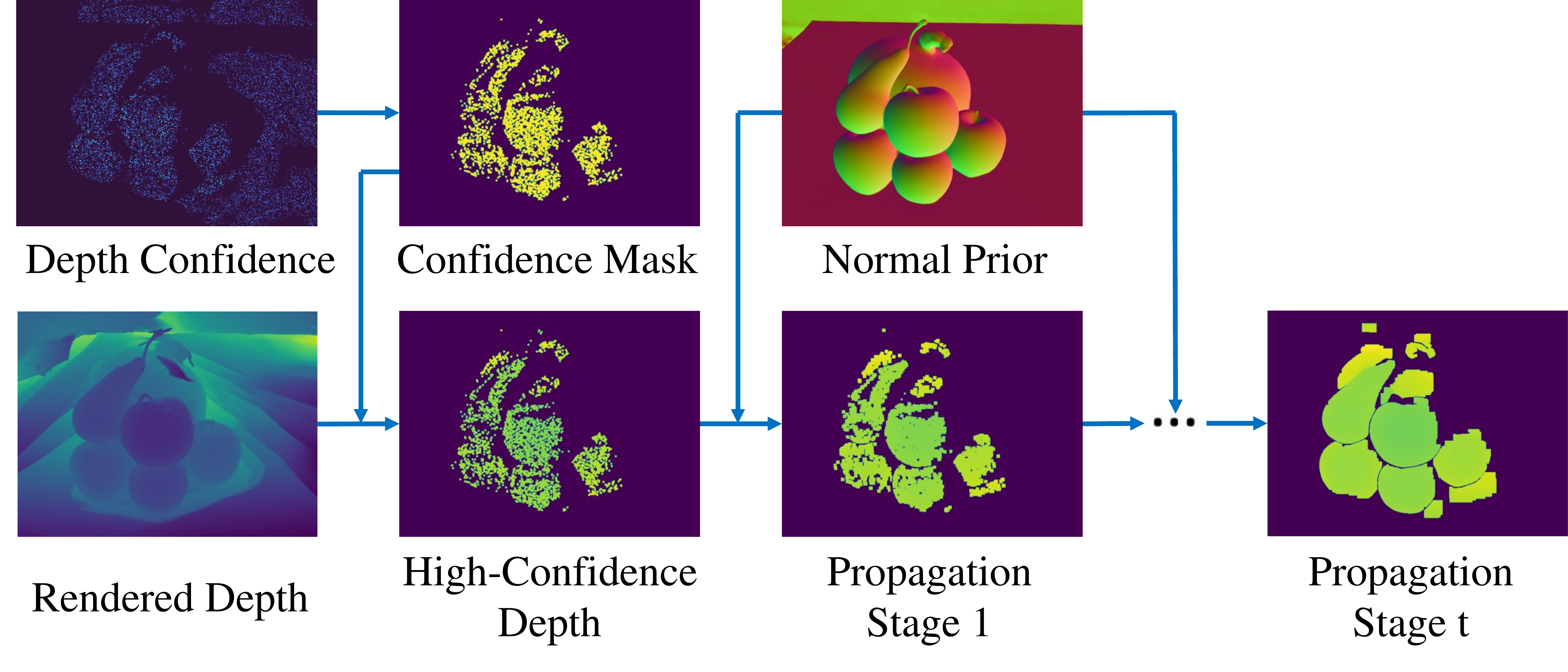}
    \caption{Ilustration of the depth propagation process.}
    \label{fig:depth_prop_process}
\end{figure}

Considering a local neighborhood around pixel $p_i$, we propagate the depth values from neighboring pixels to $p_i$. We filter the depth values through the depth confidence mask and then do the average to determine the depth of $p_i$. This process can be iterated multiple times, as shown in Fig.~\ref{fig:depth_prop_process}. Therefore, the updated depth map can be represented as,
\begin{equation}
    d_j = M_C(u_j,v_j)D ,
\end{equation}
\begin{equation}
    M_C = \left \{C > \tau \right \},
    \label{equ:confidence-threshold}
\end{equation}
\begin{equation}
d_i^{t+1}\longleftarrow \sum_{j\in N}^{} d_j^t,
\end{equation}
\begin{equation}
M_C^{t+1}\longleftarrow \sum_{j\in N}^{} M_C^t,
\end{equation}

\noindent where $D$ is the rendered depth, $M_C$ is the mask of depth confidence and $\tau$ is the threshold of the mask. $t$ denotes the iterative update, $N$ is the number of the nearest neighbors. After obtaining the updated depth map $d_i^{t+1}$ and mask $M_C^{t+1}$, it can be used to supervise the rendered depth, and the loss function is defined below, 
\begin{equation}
    L_{dp} = M_C^{t+1} \left \| D - d_i^{t+1} \right \| .
\end{equation}


Our method leverages normal priors to extend reliable depth supervision beyond the initially high-confidence regions, providing effective geometric constraints for under-constrained areas such as texture-less surfaces. Specifically, the confidence mask first identifies reliable depth regions as propagation anchors. Depth values are then propagated from these anchors to neighboring regions under the guidance of surface normals, which enforce local geometric consistency during propagation. Meanwhile, the confidence mask is progressively updated to include newly propagated regions, yielding an expanded supervision area. Depth supervision is finally applied within the updated confidence region.

\subsection{Abnormal Depth Edge-aware Smoothing Loss}
The depth propagation method may not extend depth over a large region accurately, due to its reliance on the local planar assumption and the error in the normal prior. In other words, it cannot impose constraints on global depth geometries. In particular, occluded or low-texture regions that are far from high-confidence depth areas cannot be effectively supervised. As a result, under the constraint of normal prior alone, these regions may exhibit generally correct normals but discontinuous depths, leading to abnormal depth edges, as shown in Fig.~\ref{fig:pipeline} (b). To address this issue, we propose a loss constraining on abnormal depth edge-aware smoothness.

\begin{table*}[tb]
  \centering
    \caption{Quantitative results of  Chamfer Distance (CD$\downarrow$) on DTU dataset with \textit{3 small-overlapping} images. The methods are divided into three categories, from top to bottom: (1) dense-view reconstruction methods related to ours, (2) the generalizable sparse-view reconstruction methods, and (3) the scene-specific sparse-view reconstruction methods. The best results are in \textit{bold}, the second best are \textit{underlined}. The $^\#$ indicates that the dense point cloud obtained by MASt3R \cite{duisterhof2024mast3r} is used as an initialization.}
  \resizebox{\textwidth}{!}{
  \begin{tabular}{@{}lccccccccccccccccc@{}}
    \toprule
    Scan ID & 24 & 37 & 40 & 55 & 63 & 65 & 69 & 83 & 97 & 105 & 106 & 110 & 114 & 118 & 122 & Mean CD $\downarrow$ \\
    \midrule
    2DGS \cite{huang20242dgs} & 3.54 & 4.13 & 3.61 & 1.00 & 2.69 & 2.35 & 2.04 & 2.06 & 2.94 & 1.76 & 2.40 & 2.97 & 1.35 & 2.17 & 1.69 & 2.45 \\
    PGSR \cite{pgsr} & 4.01 & 5.19 & 3.65 & 0.93 & 2.96 & 2.84 & 1.62 & 2.16 & 3.24 & 1.42 & 2.35 & 1.91 & 0.57 & 1.55 & 1.26 & 2.38 \\
    \midrule
    SparseNeus$_{ft}$ \cite{long2022sparseneus} & 4.81 & 5.56 & 5.81 & 2.68 & 3.30 & 3.88 & 2.39 & 2.91 & 3.08 & 2.33 & 2.64 & 3.12 & 1.74 & 3.55 & 2.31 & 3.34 \\
    VolRecon \cite{ren2023volrecon} & 3.05 & 4.45 & 3.36 & 3.09 & 2.78 & 3.68 & 3.01 & 2.87 & 3.07 & 2.55 & 3.07 & 2.77 & 1.59 & 3.44 & 2.51 & 3.02 \\
    GenS$_{ft}$ \cite{peng2023gens} & 7.67 & 8.58 & 5.41 & 5.88 & 7.18 & 5.26 & 4.40 & 6.29 & 6.19 & 4.08 & 6.04 & 5.29 & 4.69 & 4.35 & 4.01 & 5.69 \\
    ReTR \cite{liang2024retr} & 3.30 & 3.60 & 3.37 & 2.85 & 2.91 & 3.07 & 2.26 & 2.29 & 2.25 & 2.05 & 2.80 & 2.80 & 1.52 & 2.29 & 2.11 & 2.63 \\
    UFORecon \cite{na2024uforecon} & 1.51 & \underline{2.58} & 1.82 & 1.44 & 1.60 & 1.81 & 1.04 & 1.56 & 0.96 & 1.40 & 1.20 & 0.93 & 0.66 & 1.26 & 1.26 & 1.40 \\
    \midrule
    MonoSDF \cite{yu2022monosdf} & 3.47 & 3.61 & 2.10 & 1.05 & 2.37 & 1.38 & 1.41 & 1.85 & 1.74 & 1.10 & 1.46 & 2.28 & 1.25 & 1.44 & 1.45 & 1.86 \\
    NeuSurf \cite{huang2023neusurf} & 1.35 & 3.25 & 2.50 & 0.80 & 1.21 & 2.35 & 0.77 & 1.19 & 1.20 & 1.05 & 1.05 & 1.21 & \textbf{0.41} & 0.80 & 1.08 & 1.35 \\
    SparseCraft \cite{sparsecraft} & 2.42 & 2.79 & 2.78 & 0.74 & 1.44 & 2.51 & 1.26 & 1.42 & 1.65 & 1.10 & 1.34 & 5.24 & 0.65 & 0.88 & 1.16 & 1.83 \\
    SparseRecon \cite{han2025sparserecon} & 1.26 & \textbf{1.46} & \underline{1.39} & \textbf{0.71} & 1.20 & 2.38 & \underline{0.70} & 1.23 & 0.92 & 0.80 & \underline{0.94} & 0.77 & \underline{0.44} & 0.83 & 0.91 & 1.06 \\
    FatesGS \cite{fatesgs} & 1.32 & 2.85 & 2.71 & 0.80 & 1.44 & 2.08 & 1.11 & 1.19 & 1.33 & \underline{0.76} & 1.49 & 0.85 & 0.47 & 1.05 & 1.06 & 1.37 \\
    Sparse2DGS \cite{wu2025sparse2dgs} & 3.13 & 5.82 & 2.98 & 1.16 & 4.38 & 2.35 & 1.32 & 4.26 & 1.66 & 1.09 & 1.78 & 1.89 & 0.66 & 1.29 & 1.22 & 2.33 \\
    MAtCha \cite{guedon2025matcha} & 1.38 & 2.95 & 1.70 & 1.01 & 1.43 & 1.58 & 1.15 & 1.65 & 1.81 & 1.35 & 1.39 & 1.24 & 0.89 & 1.47 & 1.20 & 1.48 \\
    MAtCha$^\#$ \cite{guedon2025matcha} & \underline{0.88} & 2.79 & 1.40 & 0.83 & 0.93 & 1.35 & 0.87 & 1.25 & 1.51 & 0.95 & 1.15 & 1.02 & 0.56 & 1.01 & 0.92 & 1.16 \\
    Ours & 1.45 & 2.21 & 2.04 & \underline{0.72} & \underline{0.85} & \textbf{1.09} & \textbf{0.64} & \underline{1.18} & \underline{0.76} & 0.77 & 0.98 & \textbf{0.61} & 0.46 & \underline{0.79} & \underline{0.77} & \underline{1.02} \\
    Ours$^\#$ & \textbf{0.57} & \underline{1.71} & \textbf{1.37} & \textbf{0.71} & \textbf{0.84} & \underline{1.11} & \textbf{0.64} & \textbf{1.04} & \textbf{0.70} & \textbf{0.74} & \textbf{0.76} & \underline{0.65} & 0.45 & \textbf{0.66} & \textbf{0.76} & \textbf{0.85} \\
    \bottomrule
  \end{tabular}
  }
  \label{tab:dtu_cd}
\end{table*}

To extract abnormal depth edges from all depth edges, we first obtain the edge mask of the normal map and the edge mask of the depth, respectively. Then, we multiply the depth edge mask with the inverted edge mask of the normal prior to obtain the mask of abnormal depth edges. Once the edges of the abnormal depth are identified, we apply a smoothing constraint to eliminate these anomalies. Fig.~\ref{fig:pipeline} (b) illustrates the specific process. First, we sample at the edge of abnormal depth, obtaining depth $d_i$. Then, for each sampled point, we randomly sample another point within its neighborhood, obtaining depth $d_j$. Finally, we back-project these sampled points into 3D space using camera parameters and their depths, resulting in 3D points $a_i$ and $a_j$, and form a vector $\boldsymbol{v}_{ij}$. According to the local planar assumption, the vector $\boldsymbol{v}_{ij}$ should be perpendicular to the normal $\boldsymbol{n}_i$ at the sampled point $i$. Therefore, the smoothing loss is defined as
\begin{equation}
    L_{ds} = \frac{1}{N}\sum_{i,j\in N} \boldsymbol{n}_i \boldsymbol{v}_{ij} ,
\end{equation}
\begin{equation}
    \boldsymbol{v}_{ij} = a_j - a_i .
\end{equation}

\subsection{Training Loss}
Our loss function consists of depth propagation constraint loss $L_{dp}$, abnormal depth edge smoothing loss $L_{ds}$, normal prior loss $L_{normal}$, and the commonly-used rendering error $L_{rgb}$. We define the overall loss function as,
\begin{equation}
    L = L_{rgb} + \lambda_1 L_{dp} + \lambda_2 L_{ds} + \lambda_3 L_{normal} .
    \label{equ:loss-func}
\end{equation}

\noindent where $\lambda_1$, $\lambda_2$, and $\lambda_3$ are balance weights, the normal prior loss $L_{normal}$ is computed as the $\ell_1$ distance between the rendered normal and the normal prior.


\begin{figure*}[t]
  \centering
   \includegraphics[width=1.0\textwidth]{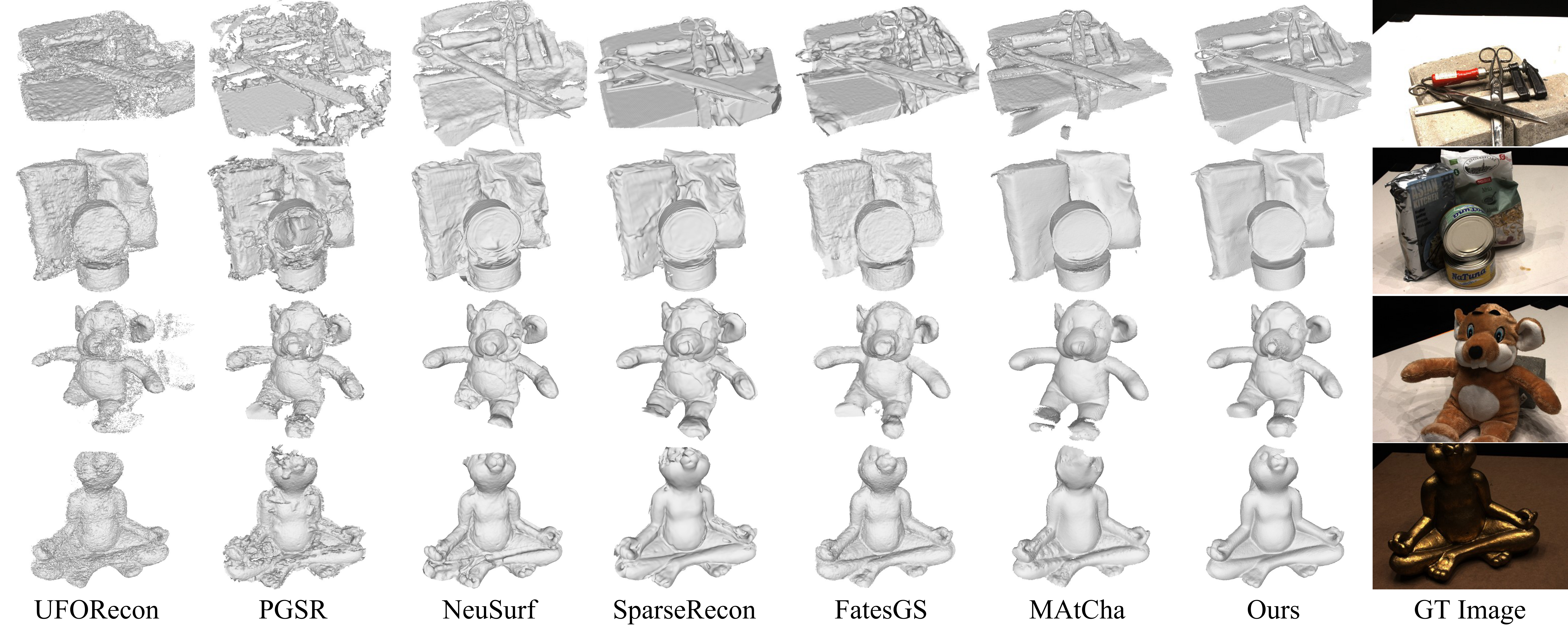}
   \caption{Visual comparison of reconstruction with 3 small-overlapping views on the DTU dataset. }
   \label{fig:dtu_visual}
\end{figure*}

\section{Experiments}
\subsection{Datasets}
We evaluate the performance of our method on three datasets: DTU \cite{jensen2014dtu}, Tanks and Temples (TNT) \cite{knapitsch2017tnt} and 5 scenes captured by ourselves. We follow previous methods~\cite{yu2022monosdf, huang2023neusurf, fatesgs, han2025sparserecon}, select 15 scenes from DTU and use 3 small-overlaping views as input. The results on views with large overlap are provided in the supplementary material.
On the TNT dataset, we follow MAtCha \cite{guedon2025matcha} and use four scenes, excluding the \textit{Courthouse} and \textit{Meetingroom} due to their large scale, which is not suitable for sparse-view reconstruction. We uniformly sample 5, 10, or 20 views from each scene as input.
We also capture 5 real-world scenes using a consumer camera. Similarly to the DTU dataset, only 3 images are used in each scene. 




\begin{figure}[t]
  \centering
   \includegraphics[width=0.8\linewidth]{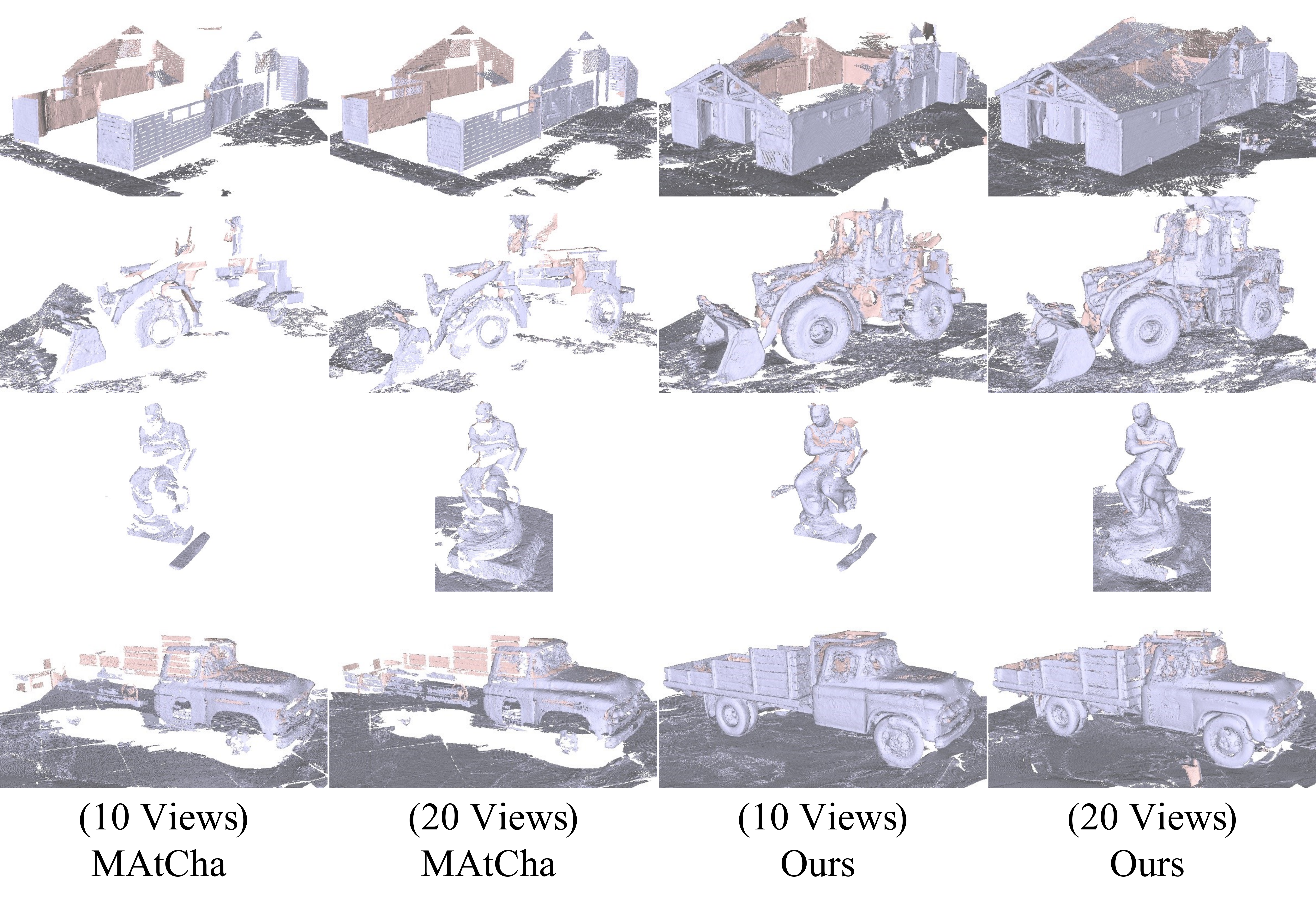}
   \caption{Visual comparison of reconstruction with sparse views on the TNT dataset.}
   \label{fig:tnt_compare}
\end{figure}

\subsection{Implementation Details}
We implement our method based on the 3DGS \cite{3dgs} framework and use the modified differentiable Gaussian renderer from PGSR \cite{pgsr}. The Gaussian opacity decay strategy \cite{han2024binocular} is also used to reduce redundant Gaussians near the scene surface. We train our model on an RTX3090 GPU, the training iterations are set to 10K.
We adopt Metric3Dv2 \cite{hu2024metric3d} to generate the normal prior. The confidence threshold $\tau$ in \cref{equ:confidence-threshold} is set to 0.6. The iterations $t$ for depth propagation is set to 10. The hyper-parameters in \cref{equ:loss-func} are set as $\lambda_1=0.5$, $\lambda_2=0.03$, $\lambda_3=0.1$.

\subsection{Baselines and Metrics}
We compare our approach with the latest sparse-view reconstruction methods including the generalizable methods, such as SparseNeuS \cite{long2022sparseneus}, VolRecon \cite{ren2023volrecon}, GenS \cite{peng2023gens}, ReTR \cite{liang2024retr} and UFORecon \cite{na2024uforecon}, and the scene-specific optimization methods, such as MonoSDF \cite{yu2022monosdf}, NeusSurf \cite{huang2023neusurf}, SparseCraft \cite{sparsecraft}, SparseRecon \cite{han2025sparserecon}, MAtCha \cite{guedon2025matcha} and FatesGS \cite{fatesgs}. We also compare our approach with the latest dense-view reconstruction methods based on 3DGS, such as 2DGS \cite{huang20242dgs} and PGSR \cite{pgsr}. 
To evaluate the reconstruction quality, we report the Chamfer Distance (CD) on the DTU dataset and F1 score on TNT dataset.

\subsection{Comparisons}

\paragraph{Sparse View Reconstruction on DTU.} 
Tab.~\ref{tab:dtu_cd} presents the CD results on the DTU dataset using 3 small-overlapping views. For a fair comparison, we also evaluate the performance of MAtCha \cite{guedon2025matcha} using COLMAP \cite{schoenberger2016colmap} point clouds for initialization. Quantitatively, our method outperforms all baseline approaches regardless of the initialization used. 
Fig.~\ref{fig:dtu_visual} provides a visual comparison of the reconstruction results. The latest generalizable methods UFORecon \cite{na2024uforecon} produces overly coarse sufaces. PGSR \cite{pgsr}, the state-of-the-art dense-view reconstruction method based on 3DGS \cite{3dgs}, also struggles to achieve fine details under sparse-view settings. The latest scene-specific optimization methods, NeuSurf \cite{huang2023neusurf}, SparseRecon \cite{han2025sparserecon}, FatesGS \cite{fatesgs} and MAtCha \cite{guedon2025matcha}, successfully reconstruct complete geometric surfaces but exhibit noticeable shortcomings.
Among all methods, our approach achieves smoother and more accurate surfaces, significantly outperforming the others.


\begin{table}[t]
    \caption{Quantitative results of F1 score on TNT dataset. The dense point cloud obtained by MASt3R is used as initialization.}
    \centering
    \begin{tabular}{@{}lccc@{}}
    \toprule
         Methods & 5 Views & 10 Views & 20 views \\
    \midrule
         FatesGS \cite{fatesgs} & 0.014 & 0.025 & 0.033 \\
         MAtCha \cite{guedon2025matcha} & 0.072 & 0.156 & 0.218 \\
         Ours & \textbf{0.081} & \textbf{0.173} & \textbf{0.293} \\
    \bottomrule
    \end{tabular}
    \label{tab:tnt_F1}
\end{table}

\paragraph{Sparse View Reconstruction on TNT.}
We compare several latest scene-specific optimized 3DGS-based sparse-view reconstruction methods on 360-degree surround-view scenes from the TNT \cite{knapitsch2017tnt} dataset. The evaluated methods include FatesGS \cite{fatesgs}, MAtCha \cite{guedon2025matcha} and ours. 
Tab.~\ref{tab:tnt_F1} show quantitative comparisons on the TNT dataset with 5, 10 or 20 views. 
FatesGS fails to reconstruct in almost all scenes and under various input conditions. Our method outperforms all 3DGS-based sparse view reconstruction methods.
Fig.~\ref{fig:tnt_compare} provides a visual comparison of reconstruction results with 10 or 20 views. MAtCha exhibits suboptimal performance and struggles to recover meaningful meshes. In contrast, our method achieves more complete and detailed surfaces. 


\begin{table}[t]
\centering
\caption{Efficiency comparison of sparse view construction methods. The GPU memory usage is recorded during training.}
\begin{tabular}{lcc}
\toprule
Methods & GPU Mem. Usage & Training Time \\
   \midrule
   UFORecon \cite{na2024uforecon} & 23 GB & 10 days \\
   SparseRecon \cite{han2025sparserecon} & 8 GB & 3.5 hours \\
   FatesGS \cite{fatesgs} & 4 GB & 14 mins \\
   MAtCha \cite{guedon2025matcha} & \textbf{3 GB} & \textbf{10 mins} \\
   Ours      & \textbf{3 GB}             & \textbf{10 mins}      \\
   \bottomrule
\end{tabular}
\label{tab:efficiency_compare}
\end{table}

\begin{figure}[t]
  \centering
   \includegraphics[width=1.0\linewidth]{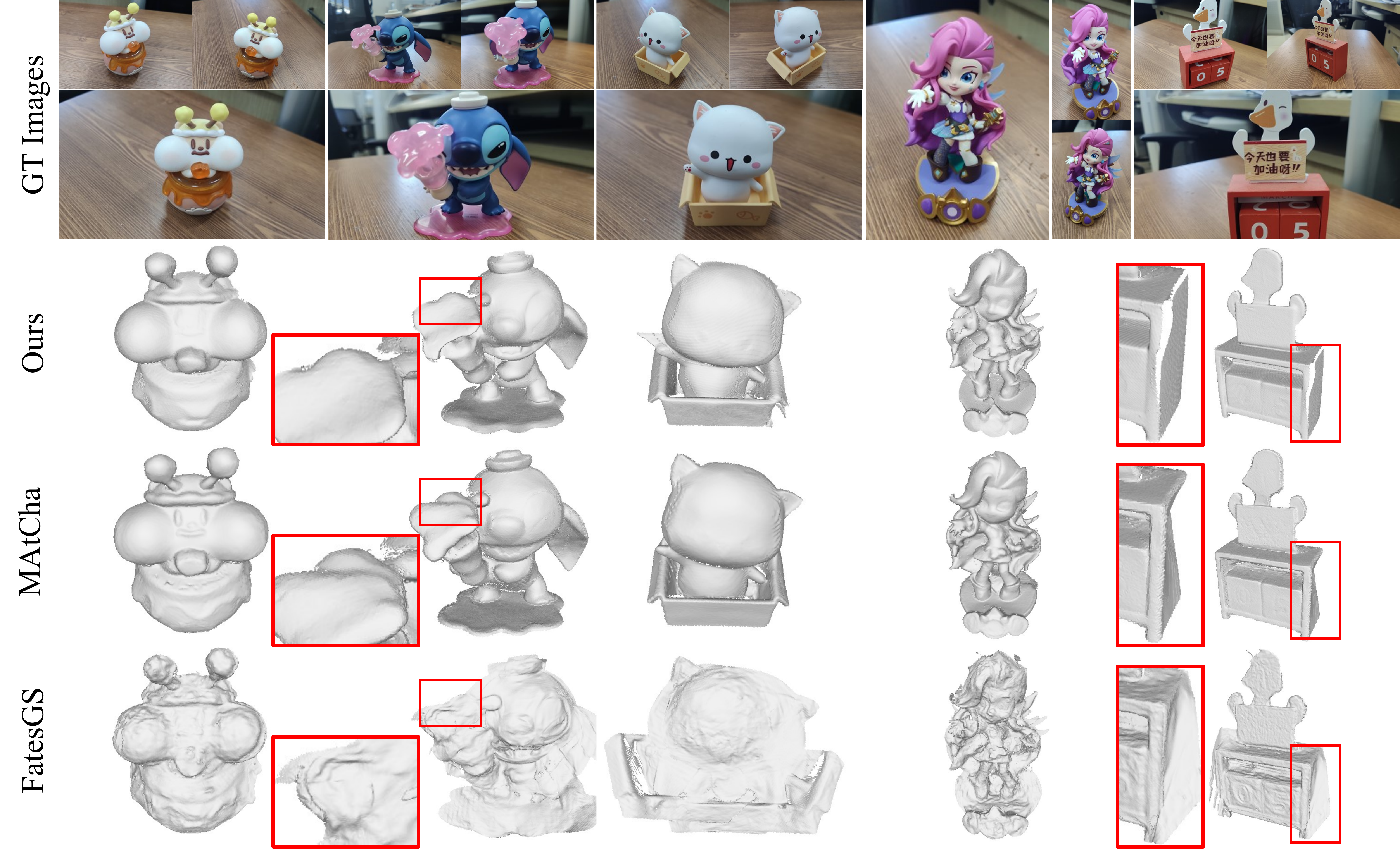}
   \caption{Visual comparison on self-captured dataset with 3 input views.}
   \label{fig:selfdata_compare}
\end{figure}

\paragraph{Sparse View Reconstruction on Self-Captured Data.}
On our self-captured dataset, we compare our method with FatesGS \cite{fatesgs} and MAtCha \cite{guedon2025matcha}, as these two methods, specifically designed for scene-specific sparse-view reconstruction, are more comparable to ours in terms of both reconstruction quality and efficiency. As shown in Figure \ref{fig:selfdata_compare}, 
the results of MAtCha exhibit some shortcomings, while the results of FatesGS show noticeable distortions and roughness. In general, our method achieves the best reconstruction performance.

\paragraph{Efficiency Comparison.} 
Although generalizable methods can produce reconstructions in minutes, they require extensive pre-training, often spanning several days, and demand more GPU memory. Scene-specific methods based on NeRF involve time-consuming optimization. In contrast, 3DGS-based approaches have demonstrated much higher training efficiency. As shown in Tab.~\ref{tab:efficiency_compare}, our method achieves comparable memory usage and training time to other 3DGS-based optimization frameworks. Notably, our approach attains superior reconstruction quality under comparable computational efficiency.

\subsection{Ablation Study}

\begin{wraptable}{r}{0.5\linewidth}
  \centering
    \caption{Ablation studies with 3 small-overlapping views on DTU dataset. O.D. stands for the opacity decay strategy.}
  \begin{tabular}{@{}ccccc@{}}
    \toprule
    $L_{normal}$ & O.D. & $L_{dp}$ & $L_{ds}$ & Mean CD$\downarrow$\\
    \midrule
     & & & & 2.79\\
    \checkmark & & & & 2.28 \\
    \checkmark & \checkmark & & & 1.55 \\
    \checkmark & \checkmark & \checkmark & & 1.13 \\
    \checkmark & \checkmark &  & \checkmark & 1.38 \\
    \checkmark & \checkmark & \checkmark & \checkmark & \textbf{1.02} \\
    \bottomrule
  \end{tabular}
  \label{tab:ablation}
\end{wraptable}

To validate the effectiveness of the depth propagation constraint loss $L_{dp}$ and the anomalous depth edge smoothing loss $L_{ds}$, we conduct an ablation study on the DTU \cite{jensen2014dtu} dataset and report the average Chamfer Distance (CD) across 15 scenes. As shown in Tab.~\ref{tab:ablation}, removing any of our proposed optimization losses results in performance degradation, showing the effectiveness of each component. Since the anomalous depth edges only appear in a small subset of scenes in the DTU dataset, the performance improvement from the anomalous depth edge smoothing loss is relatively limited.

\begin{wrapfigure}{r}{0.5\linewidth}
    \centering
   \includegraphics[width=1.0\linewidth]{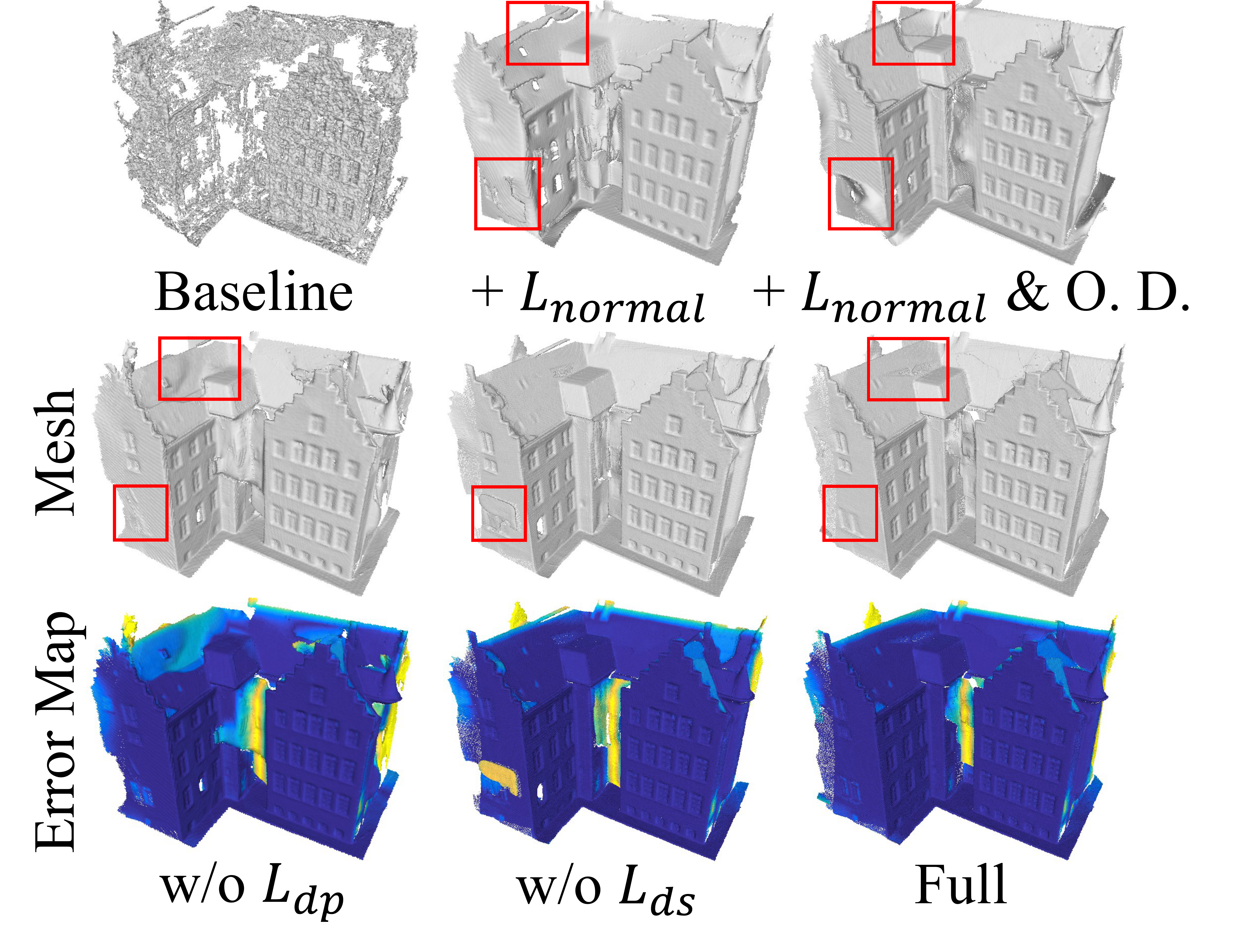}
   \caption{Visualization of ablation results on \textit{Scan 24} in the DTU dataset. O.D. stands for the opacity decay strategy.}
   \label{fig:ablation_visual}
\end{wrapfigure}

Fig.~\ref{fig:ablation_visual} presents the visualization of ablation results on scan 24 of the DTU dataset. We use a model that incorporates only the rendering loss as the baseline.
Starting from this baseline, we progressively add the normal loss $L_{normal}$ and the opacity decay strategy \cite{han2024binocular}. The first row of Fig.~\ref{fig:ablation_visual} shows the corresponding reconstruction results. As observed, the loss $L_{normal}$ smooths the reconstructed surface, but distortions remain. When the opacity decay strategy is further introduced, these distortions are alleviated to some extent, but noticeable shortcomings persist.

To verify the contribution of either the depth propagation constraint loss $L_{dp}$ or the depth edge smoothing loss $L_{ds}$, we add each of them to the normal loss and the opacity decay strategy. The second and third rows in Fig.~\ref{fig:ablation_visual} illustrate the corresponding reconstruction results and error maps respectively.
It can be seen that without $L_{ds}$, the surface is not complete, whereas omitting $L_{dp}$ leads to slight deformations. These deformed regions, characterized by minimal depth variations, are difficult to detect through depth edge, making $L_{ds}$ ineffective in addressing them. However, the depth propagation constraint loss $L_{dp}$ effectively eliminates these distortions.

\begin{wrapfigure}{r}{0.5\linewidth}
    \centering
    \includegraphics[width=1.0\linewidth]{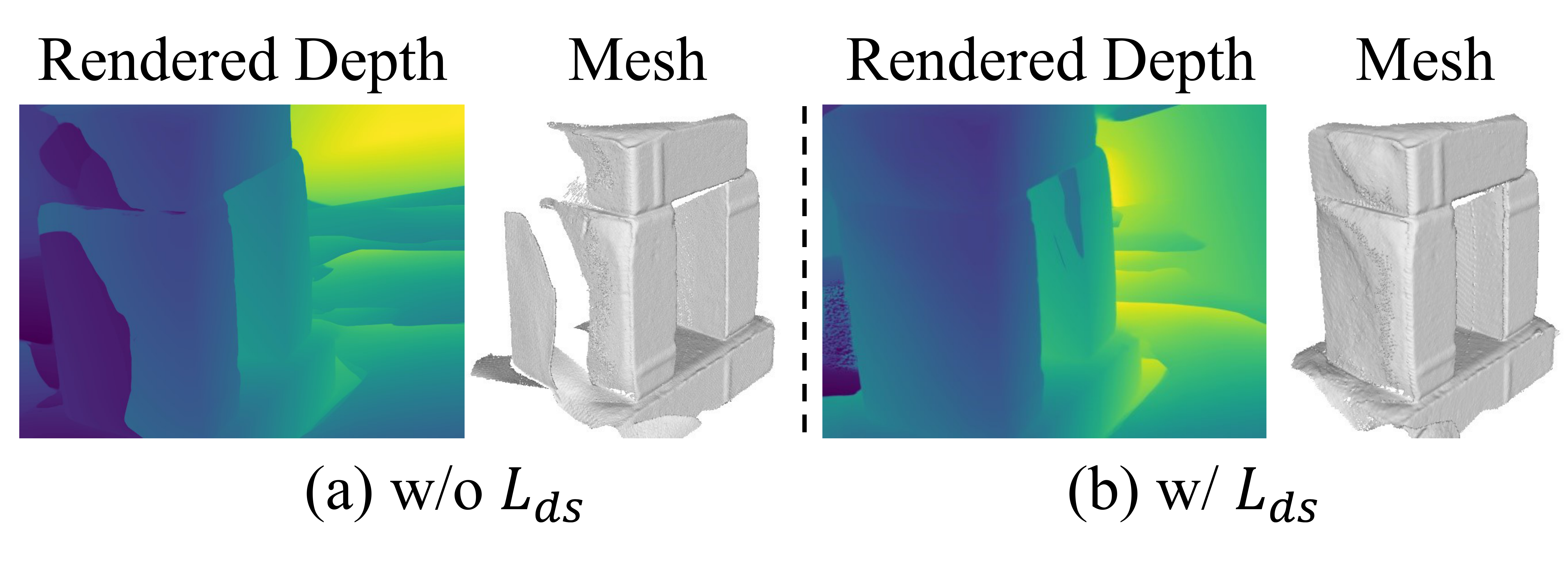}
    \caption{Effect of the abnormal depth edge-aware smoothing loss.}
    \label{fig:depth_smooth_effect}
\end{wrapfigure}

Fig.~\ref{fig:depth_smooth_effect} provides a clearer illustration of the depth anomalies, which lead to mesh fragmentation. The comparison shows that $L_{ds}$ significantly improves the quality of the reconstructed geometry.


To investigate the impact of the iterations $t$ of depth propagation on reconstruction quality, we conduct an ablation on the DTU dataset. As shown in Tab.~\ref{tab:ablation_prop_iters}, when the iterations $t$ increases from 4 to 10, the CD consistently decreases, indicating a steady improvement in reconstruction accuracy. However, further increasing the iterations beyond 10 does not yield additional gains and even leads to slight fluctuations in CD values. Therefore, we set the number of propagation iterations to 10 in our implementation. Notably, even when $t$ is increased to 20, the total computation time remains as low as 0.008 seconds, having almost no impact on the overall performance of our method.

\begin{table}
    \centering
        \caption{Effect of propagation iterations on Chamfer Distance (CD) on the DTU dataset.}
    \begin{tabular}{cccccccccc}
         \toprule
         $t$ & 4 & 6 & 8 & 10 & 12 & 14 & 16 & 18 & 20 \\
         \midrule
         CD $\downarrow$ & 1.126 & 1.069 & 1.052 & 1.021 & 1.063 & 1.027 & 1.019 & 1.056 & 1.032 \\
         \bottomrule
    \end{tabular}
    \label{tab:ablation_prop_iters}
\end{table}

To evaluate the robustness of our method to the quality of normal priors, we conducted an ablation study on the DTU dataset using three different normal estimators: Omnidata \cite{eftekhar2021omnidata}, StableNormal \cite{ye2024stablenormal} and Metric3Dv2 \cite{hu2024metric3d}. As shown in Tab.~\ref{tab:ablation_diff_normals}, higher-quality normals leads to lower CD. However, it is worth noting that our method still achieves reconstruction quality comparable to the recent MAtCha \cite{guedon2025matcha} method even when using the relatively low-quality normals from Omnidata \cite{eftekhar2021omnidata}. This demonstrates that our framework is robust to the accuracy of the normal prior.

\begin{table}
    \centering
        \caption{Effect of different normal priors on reconstruction quality on the DTU dataset.}
    \begin{tabular}{lccc}
        \toprule
         Normal Prior &  Omnidata \cite{eftekhar2021omnidata} & Metric3Dv2 \cite{hu2024metric3d} & StableNormal \cite{ye2024stablenormal} \\
         \midrule
         CD $\downarrow$ & 1.14 & 1.02 & 1.05  \\
         \bottomrule
    \end{tabular}
    \label{tab:ablation_diff_normals}
\end{table}

\section{Conclusion}
In this paper, we propose a novel 3DGS-based surface reconstruction method, achieving fast and high-fidelity geometric reconstruction from sparse views. We leverage normal priors to guide the propagation of high-confidence depth regions, thereby constraining areas with lower depth confidence. Additionally, we perform anomalous depth edge detection to address depth discontinuities. These constraints ensure the accuracy of rendered depth under sparse input views, ultimately leading to high-quality scene geometry. Extensive experiments on DTU, TNT, and self-captured datasets demonstrate that our method outperforms state-of-the-art sparse view surface reconstruction methods.

\paragraph{Limitation.}
In our method, the depth confidence is estimated based on multi-view geometric and photometric consistency. However, these consistency cues can become unreliable in texture-less regions, leading to inaccurate confidence estimation. As a result, erroneous depths may be propagated to neighboring areas, resulting in local geometric distortions or artifacts in the reconstructed surface.

\section*{Acknowledgements}
The corresponding authors are Yu-Shen Liu and Junsheng Zhou. This work was partially supported by Deep Earth Probe and Mineral Resources Exploration -- National Science and Technology Major Project (2024ZD1003405), and the National Natural Science Foundation of China (62272263).

%
%
\bibliographystyle{splncs04}
\bibliography{main}
\end{document}